\documentclass[11pt]{article}
\usepackage{acl2014}
\usepackage{times}
\usepackage{url}
\usepackage{latexsym}
\usepackage{amsmath,amsfonts}
\usepackage{graphicx}
\usepackage{enumitem}

\title{Rule-based Emotion Detection on Social Media:\\ Putting Tweets on Plutchik's Wheel}

\author{Erik Tromp$^{1,2}$ \\
  $^{1}$Adversitement B.V. \\
  The Netherlands \\
  {\tt erik.tromp@adversitement.com} \\\And
  Mykola Pechenizkiy$^{2}$ \\
  $^{2}$Dept. of Computer Science, TU Eindhoven \\
  The Netherlands \\
  {\tt m.pechenizkiy@tue.nl} \\}

\date{}

\begin{document}
\maketitle

\begin{abstract}
We study sentiment analysis beyond the typical granularity of polarity and instead use Plutchik's wheel of emotions model.
We introduce RBEM-Emo as an extension to the Rule-Based Emission Model algorithm to deduce such emotions from human-written messages.
We evaluate our approach on two different datasets and compare its performance with the current state-of-the-art techniques for emotion detection, including a recursive auto-encoder. The results of the experimental study suggest that RBEM-Emo is a promising approach advancing the current state-of-the-art in emotion detection.
\end{abstract}

\section{Introduction}
\label{sec:introduction}

Current sentiment analysis methods - ranging from baseline bag-of-words methods to state-of-the-art neural methods - typically focus on deducing information on subjectivity or polarity only (Section~\ref{sec:relatedwork}).  
Human emotions move far beyond these simple metrics and are much more diverse. This implies that such subjectivity- or polarity-analysis only gives limited information on the actual intent of an author of a message.

Defining axes of polarity is not a hard task, typically one has negativity, positivity and a notion of neutrality or objectivity in between. For emotions however, defining a complete and clear set of emotions is much more difficult. Though several researchers attempted at defining standards in this field~\cite{parrott:2001,plutchik:1980,schroder:2011}, AAAC\footnote{The Association for the Advancement of Affective Computing - http://emotion-research.net/}, there is still no consensus on a basic set of emotions that is generally accepted and could be objectively verified.

The goal of this paper is to present a sentiment analysis approach accompanied by a model of emotions that fit well together in order to set a standard in emotion analysis to expand upon.






We present a new \emph{RBEM-Emo} approach for emotion detection from human-written texts.\footnote{We expect a revised and extended version of this manuscript describing RBEM-Emo to appear in~\cite{Tromp2015chapter} } 
This algorithm is based on work by~\cite{tromp:2013} where the authors introduced the Rule-Based Emission Model (RBEM) algorithm for polarity detection only.
RBEM generates positive and negative emissions based on several groups of patterns that capture various ways how sentiment can be expressed in natural language.
We show how this approach can be developed further to go beyond polarity and measure emotions as given by Plutchik's wheel of emotions.


We conducted an experimental evaluation of RBEM-Emo on a publicly available benchmark and on a new benchmark that we constructed. 
The results of our evaluation suggest that RBEM-Emo outperforms the current state-of-the-art approaches for emotion detection. To facilitate reproducibility of the results and further progress in emotion classification 
we made our benchmark publicly available.

\section{Related Work}
\label{sec:relatedwork}


Moving beyond polarity in sentiment analysis in currently upcoming and not well studied yet. Few examples can be found where novel methods are introduced to capture more information than just polarity such as the work of \cite{socher:2011} where a recursive auto-encoder is used to predict sentiment distributions in five dimensions. \cite{cambria:2012a} and~\cite{cambria:2012b} promote affective computing using a framework they call SenticNet. The sentiment dimensions of this framework are modeled in an hourglass-model which is a derivative of Plutchik's wheel of emotions \cite{plutchik:1980}. In~\cite{mohammad:2012:STARSEM-SEMEVAL} the author collected and experimented with a large collection of tweets with self-labeled emotion hashtags.

The closest work to our approach is~\cite{Andreevskaia:2007}, in which the authors considered a rule-based approach based on a set of positive and negative patterns and valence shifters for handling negations and other linguistic constructs defining the sentiment of a sentence.

Standards on emotion frameworks are difficult to define as emotions are usually subjective and cannot be crisply defined. Works of \cite{parrott:2001,plutchik:1980,schroder:2011} do aim to define standards in this area by defining a minimal set of basic emotions from which more complex ones can be derived or constructed by combining basic emotions. In \cite{cambria:2012b} the authors develop methods to reason about emotions.
In \cite{ekman:1989}, facial expressions are linked to emotions and a final six universal basic emotions are presented.


\section{Approach to Emotion Detection}
\label{sec:approach}

\subsection{Plutchik's Wheel of Emotions}

\begin{figure}[!htbp]
    \centering
    \includegraphics[width=0.95\columnwidth]{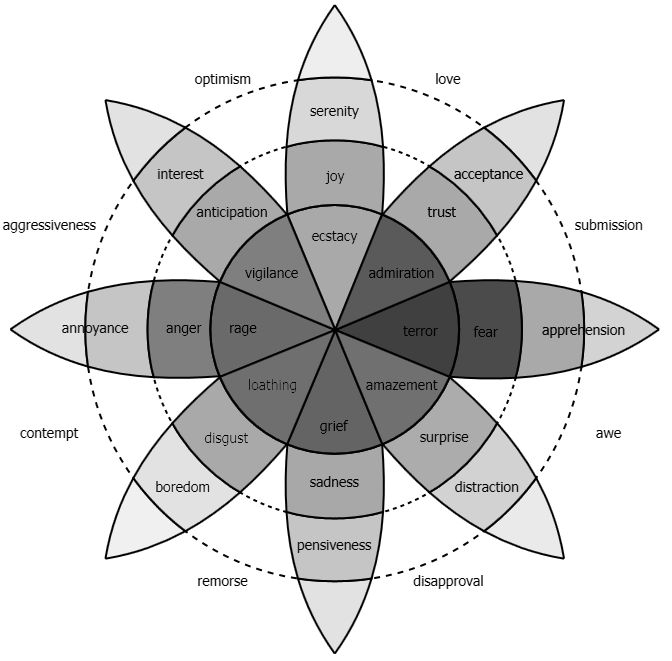}
    \caption{Plutchik's wheel of emotions \cite{plutchik:1980}}
    \label{fig:plutchik_wheel}
\end{figure}

To tackle the problem of emotion detection, one needs to have a notion of emotion. As e.g.\ in text mining the problem can be formulated differently depending on whether we have just two classes like in spam filtering, or several categories like topic classification or a large number of categories like in automated tagging. 
We choose the wheel of emotions defined by Robert Plutchik~\cite{plutchik:1980} (see Figure~\ref{fig:plutchik_wheel}) because it defines only eight basic emotions, which makes the problem manageable for envisioned applications and RBEM-Emo a good match to perform classification according to this model of emotions.

\label{sec:approach:plutchik}

These eight emotions are assumed to be complete in the sense that any expressed emotion is related or subsumed by one of the eight. In his work, Plutchik states that these emotions are culturally independent. Given this assumption, we can apply this model to any given language, which we consider to be a strong point.

Another reason for using this model is that each of these eight basic emotions are opposites of one of the other basic emotions. This means that we can in fact measure four axes where opposite emotions exist on the two extremes of a single axis. Additionally, Plutchik defines eight \emph{human feelings} that are derivatives of combinations of two basic emotions. This in fact means that with modeling only four axes, we can get a total of sixteen dimensions of emotions and feelings.

\subsection{RBEM for Emotion Detection}
\label{sec:approach:rbem}
In our previous work we conducted several case studies with RBEM illustrating that it is rather generic and easily extendable allowing to develop different solutions that are scalable, transparent, and easy to maintain and adapt for the needs of a particular domain. We considered integrating RBEM into a larger data analytics project~\cite{DBLP:conf/icdm/TrompP11} and into mobile settings with computing resources be a bottleneck~\cite{DBLP:conf/kes/ChambersTPG12}. Hence, we had a good incentive to extend it to emotion classification on social media.

The original Rule-Based Emission Model (RBEM) algorithm \cite{tromp:2013} can be used for polarity detection assigning new messages a label that is one of \emph{positive, neutral, negative}.
The algorithm's internals work in such a way that either positive or negative emissions can be generated upon which subsequently different rules are executed to modify these emissions.

The rules work on patterns that belong to one of the following groups:
{positive} and {negative} patterns, e.g.\ \emph{good, well done} and \emph{bad, terrible};
{amplifier} and {attenuator} patterns to strengthen or weaken polarity of entities \emph{very much, a lot} and \emph{a little, a tiny bit}; right- and left-flip patterns to handle negations, e.g.\ \emph{not, no} and sentences containing constructs with e.g.\ \emph{but, however}; {continuator} patterns to handle constructs with e.g.\ \emph{and, and also}; {stop} patterns to interrupt the emission of polarity when punctuation signs such as a dot or an exclamation mark, expressing the general case that polarity does not cross sentence boundaries, appear in a message.


\begin{table*}
    \noindent
    {\bf Emission of emotions rules:}
    \begin{align*}
    \forall_{\mathit{emo} \in \{Joy, Sadness, Trust, Disgust, Fear, Anger, Surprise, Anticipation\}}: \\
      \forall_{(s, f, \mathit{emo}) \in \mathit{maxPatterns}}:c = \lfloor\frac{s + f}{2}\rfloor \wedge  
      (\forall_{e_i \in m}: \neg(\exists_{t \in \mathit{stops}}:c \geq i \Rightarrow i \leq t \leq c \vee 
      i \geq c \Rightarrow c \leq t \leq i) \\
      \Leftrightarrow em_{esc_{\mathit{emo}}}(e_i) = em_{esc_{\mathit{emo}}}(e_i) + sign_{\mathit{emo}} \cdot e^{-i} )
    \end{align*}
\end{table*}

Crucial to the algorithm is that positivity and negativity are opposites of each other and hence allow for example negations to simply invert the emission. This specific characteristic of the algorithm makes it work well with Plutchik's model since the emotions defined in that model are also opposites of each other. We in fact extend the RBEM algorithm to perform the same type of rules but now -- instead of having one axis to measure; positive on one end of the extreme and negative on the other extreme -- we have four different axes, together yielding eight different emotions being measured.

The RBEM algorithm requires pattern groups to be defined.
It uses a pattern matching on wildcards to identify patterns in a message.
When classifying previously unseen messages, two steps are performed. First all patterns in the model that match a message are collected. Then, rule(s) associated with each pattern group for each pattern present in the message are applied.

This actual internal algorithms for constructing and applying RBEM remain unchanged. We refer to the original paper on RBEM for their formal description~\cite{tromp:2013}.

RBEM-Emo extends RBEM for emotion detection by introducing new pattern groups.
The RBEM algorithm uses two base pattern groups to define emission of polarity, positive and negative patterns. For our RBEM-Emo algorithm, we replace these two pattern groups with eight new pattern groups, one for each basic emotion of Plutchik's model: \emph{joy, sadness, trust, disgust, fear, anger, surprise, anticipation}. Similarly, we replace the two rules that are defined on positive and negative patterns with eight new rules. Note that conceptually, we perform the exact same process we do for positive polarity on one hand and negative polarity on the other hand, but now four times, once for each axis.

Since we no longer operate on a single emission score but instead on four, we define a mapping from emotions to an index by $esc_{emo}$ and we define a sign counterparts $sign_{emo}$ for each emotion on a single axis. Here $esc_{Joy} = esc_{Sadness} = 1$ and $sign_{Joy} = 1, sign_{Sadness}  = -1$, $esc_{Trust} = esc_{Disgust} = 2$ and $sign_{Trust} = 1, sign_{Disgust}  = -1$, $esc_{Fear} = esc_{Anger} = 3$ and $sign_{Fear} = 1, sign_{Anger}  = -1$, $esc_{Surprise} = esc_{Anticipation} = 4$ and $sign_{Surprise} = 1, sign_{Anticipation}  = -1$. We also define a subscripted emission score $em_j(e_i)$ where $j \in [1, 4]$ and the value of $j$ corresponds with the emotion axis for the emotions that map to $j$ using $esc_{emo}$ (i.e.\ $em_1$ is the axis function used by Joy and Sadness).

The new rules that replace the original rules defining positive sentiment emission and negative sentiment emission are defined as shown at the top of the page\footnote{Note that the RBEM algorithm requires rules to be executed in-order.}.

All the the other original RBEM rules are executed four times, once for every $em_j, j \in [1, 4]$. When the algorithm terminates, this yields us four emission scores, i.e.\ one score per dimension.

Once the algorithm has terminated, we can obtain a total score for each pair or opposite emotions, e.g.\ for Joy and Sadness by summing of all emissions of $em_j$. $\mathit{JoySadness} = \sum_{i=1}^{n} em_1(e_i)$. Whenever $\mathit{JoySadness} > 0$ we say that Joy was expressed in the original message. Similarly, when $\mathit{JoySadness} < 0$, we say that Sadness was expressed. If $\mathit{JoySadness} = 0$, neither Joy nor Sadness was expressed. The other three emission axes can be interpreted similarly.

As an illustrative example, consider the sentence \emph{I thought I would like the new XYZ phone, but now that I have it, it is a huge disappointment, it makes me angry}. Suppose also that we have the following patterns (Part-of-Speech tags left out for simplicity): $(I * like, Anticipation)$, $(but, Leftflip)$, $(huge _, Amplifier)$, $(disappointment, Sadness)$, $(angry, Anger)$. The algorithm would first assign the emotion scores to all parts of the sentence where patterns are found. This would yield the first part emitting negatively on $em_4$, the third phrase emitting negatively on $em_1$ and the last phrase emitting negatively on $em_3$. Next, the scores on pattern indicated by the word \emph{huge} will amplify the emissions on all axes, with the biggest effect on $em_1$. Finally, the leftflip indicated by \emph{but} will convert all negative emissions on its left -- influencing $em_4$ mainly -- to its opposite direction, yielding positive emissions on $em_4$. The final outcome will hence be that -- ordered by decreasing strength -- \emph{Sadness, Anger} and \emph{Surprise} are present.

\section{Experimental Evaluation}
\label{sec:experiments}
With the experimental study we aim to evaluate the proposed RBEM-Emo algorithm, which is tailored towards Plutchik's model of emotions. 

\subsection{Experiment Setup}
\label{sec:experiments:setup}
We compare our method against a majority class baseline, Support Vector Machines (SVMs), regression and the recursive auto-encoder of~\cite{socher:2011} and evaluate on accuracy. In~\cite{socher:2011} five-dimensional sentiment model originating from the Experience Project\footnote{See http://www.experienceproject.com} is introduced. It would be reasonable to evaluate on this dataset, but the five labels used to express emotions in that dataset are quite arbitrary and ambiguous\footnote{The labels are \emph{Sorry, Hugs}, \emph{You Rock}, \emph{Teehee}, \emph{I Understand} and \emph{Wow, Just Wow}}, as the authors already indicate. In addition, these labels are produced by users that read an actual confession by a different person and instead of capturing the emotion of the actual message hence capture the emotion triggered with an external reader. 

Due to the impracticalities of the Experience Project dataset for our experiments, we instead benchmark on a different, well-accepted dataset introduced in~\cite{alm:2008}. This dataset is annotated using Ekman's emotions~\cite{ekman:1989} instead of Plutchik's, but since the six basic emotions of Ekman are subsumed by the eight emotions of Plutchik's model, we can use the labels in a straightforward manner, ignoring labels produced by RBEM-Emo that do not exist in Ekman's model and producing the majority class as label in case we find a non-existing emotion. We refer to this dataset as the \emph{Affect Dataset}.

In addition to benchmarking on a well-accepted public dataset, we also introduce our own \emph{Twitter Dataset} that is annotated on Plutchik's emotions. 

For the SVM and regression classification we use LibShortText~\cite{yu:2013}. We experiment using both word counts and TF-IDF scores as features. For the recursive auto-encoder, we use the Java version referenced to by the authors of~\cite{socher:2011}\footnote{Can be found at https://github.com/sancha/jrae}. To ensure we have the right setup of the auto-encoder, we reproduced the polarity detection experiments on the rotten tomatoes dataset as done in~\cite{socher:2011} and obtained an accuracy of 77.0\%. This is in line with the results presented in~\cite{socher:2011}, illustrating our setup is valid. When we apply our RBEM-Emo classifier, we get four scores for each axis in Plutchik's model, summing up to eight emotions.
Finally, we assign a single label corresponding to the highest of all eight emotion scores.


\subsection{Datasets Description}
\label{sec:experiments:data}

The \textbf{Affect Dataset} we use is presented in~\cite{alm:2008} and is publicly available\footnote{\url{http://lrc.cornell.edu/swedish/dataset/affectdata/}}. This dataset consists of snippets of text obtained from books written by three different authors.

For each snippet, every sentence is annotated by two annotators. These annotators provide two different labels each, one for the prevailing emotion found in the sentence and one for the mood found. The available labels are the six basic emotions of Ekman's universal emotions, being \emph{angry}, \emph{disgusted}, \emph{fearful}, \emph{happy}, \emph{sad}, \emph{surprised}. In addition, the authors could also indicate neutrality. 

We use only those messages for which both annotators agree upon emotion and we discard the mood label produced by the annotators. Moreover, since 85\% of all sentences in the dataset are neutral, and many general purpose classification techniques suffer from class imbalance, we produce two different datasets, one where neutral sentences are removed and only emotion-bearing sentences are maintained and one where neutral messages are included.
For evaluation purposes, we use roughly $\frac{2}{3}$ of the data for training and $\frac{1}{3}$ for testing.
The resulting sizes of the training sets are 7527 and 1084 instances depending on the in- or exclusion of the neutral class, and for test sets -- 3590 and 488 instances correspondingly.

%

\paragraph{Twitter Dataset.} Since the proposed RBEM-Emo method is tightly integrated with Plutchik's wheel of emotions, we evaluate on data annotated on these emotions. We collected a large amount of tweets in three different languages: \emph{English}, \emph{Dutch} and \emph{German}. We had at least two independent annotators to annotate each of these messages using a dedicated Web-based annotation tool. In case of disagreement, we use the prevailing emotion label given by the annotators as actual label for a message. If there is no agreement on the prevailing emotion label, the message was discarded. 

In addition, the annotators were asked to identify patterns in these messages such that we can later on construct the RBEM-Emo model from them.

The data was collected from Twitter where a language detection algorithm was used to filter out those messages that are written in English, Dutch or German as a first step. All messages wrongly identified by language are later on filtered out by the annotators.

In line with the setup of the experiments presented in~\cite{socher:2011} and adhered to here, we randomly split the data into roughly $\frac{2}{3}$ training and $\frac{1}{3}$ test data. The resulting training/test set sizes are Dutch 289/113 for Dutch, 235/113 for English and 225/109 for German.

The Twitter dataset is made publicly available\footnote
{\url{http://www.win.tue.nl/~mpechen/projects/smm/}}.


\subsection{Results}
\label{sec:experiments:results}


The accuracies of the best performing general purpose classification techniques on the Affect Dataset are compared to those of RBEM-Emo in Table~\ref{tbl:accuracies:affect}. The majority class classification accuracy is given as a baseline. We report accuracies both for the case when neutral messages are kept in our dataset and when they are filtered out. We do this since the neutral messages compose 85\% of the entire original dataset and it is expected that generic classification techniques will suffer from class imbalance and learn biases towards this data rather than find actual emotions. This is reflected in the accuracies of the SVM and regression classifiers which are marginally higher than the majority class baseline. Surprisingly, the recursive auto-encoder (RAE) that is currently claimed to be the state-of-the-art technique for emotion classification performs worse than several simpler classifiers and in fact is as good as a majority class classifier. One possible reason for this might be that the size of our dataset is relatively small. 
RBEM-Emo classifier being a tailor approach to deduce emotional patterns outperforms the other classifiers.

In the second column of Table~\ref{tbl:accuracies:affect}, we report the accuracies when all messages belonging to the neutral class are removed, yielding a more class-balanced dataset. Here we see much better improvements over the majority class baseline for SVM and regression and now also for the recursive auto-encoder. Using TF-IDF scores for features is favored over using just word counts. The RBEM-Emo method however, still outperforms the other classifiers.

Table~\ref{tbl:accuracies:emo} lists the accuracies obtained per language on our own Twitter corpus. For each classifier, we report the accuracy on each language (being Dutch, English and German) and report a total accuracy which is the average accuracy over all messages in all three languages. A generic result over all classifiers is that the accuracies on English data seem to be the lowest, implying most ambiguity within this language. Remarkable is that the recursive auto-encoder performs worse than SVM and regression models and yields no benefit over the majority class guess. Again, this could be due to the small size of the corpus or difficulty in finding the most suitable model parameters. 
There is no clear evidence on whether TF-IDF scores or word counts work better for this dataset.
The RBEM-Emo classifiers yields the highest accuracy for each of three languages. 

\begin{table}[!htbp]
  \centering
  \begin{sc}
  \begin{small}
    \begin{tabular}{lcr}
        \hline
        Method & Acc. w/ Ntl & Acc. no Ntl\\
        \hline
        Majority & 84.4\% & 37.7\%\\
        SVM, W.C. & 86.2\% & 61.3\%\\
        SVM, TF-IDF & 86.2\% & 65.0\%\\
        Regr., W.C. & 85.8\% & 59.5\%\\
        Regr., TF-IDF & 85.5\% & 63.4\%\\
        RAE & 84.4\% & 60.4\%\\
        \textbf{RBEM-Emo} & \textbf{88.4\%} & \textbf{67.1\%}\\
        \hline
    \end{tabular}
  \end{small}
  \end{sc}
  \caption{Accuracies on the Affect dataset.}
  \label{tbl:accuracies:affect}
\end{table}

\begin{table}[!htbp]
  \centering
  \begin{sc}
  \begin{small}
      \begin{tabular}{l|ccccccc}
        \hline
        \rotatebox{90}{Language} & \rotatebox{90}{Majority} & \rotatebox{90}{SVM W.C.} & \rotatebox{90}{SVM TF-IDF} & \rotatebox{90}{Regr W.C.} & \rotatebox{90}{Regr TF-IDF} & \rotatebox{90}{RAE} & \rotatebox{90}{\textbf{RBEM-Emo}}\\
        \hline
        nl & 50.4 & 53.1 & 54.9 & 53.1 & 53.1 & 53.1 & \textbf{56.7}\\
        en & 42.5 & 46.0 & 42.5 & 45.1 & 42.5 & 31.0 & \textbf{47.2}\\
        de & 34.9 & 46.8 & 47.7 & 40.4 & 46.8 & 44.0 & \textbf{53.2}\\
        \emph{all} & 42.7 & 48.7 & 48.4 & 46.3 & 47.5 & 42.7 & \textbf{52.4}\\
        \hline
    \end{tabular}
  \end{small}
  \end{sc}
  \caption{Accuracies on the Twitter dataset.}
 \label{tbl:accuracies:emo}
\end{table}

\section{Conclusions}
\label{sec:conclusions}

In this work we have introduced a new rule-based classification technique called RBEM-Emo for emotion classification on social media. This emotion classification approach is tightly coupled with the Plutchik's model of emotions. We proposed to use this model because it relatively compact yet complete and models emotions as opposites of each other, a feature that works well with RBEM-Emo.

The results of our experimental study show that RBEM-Emo is competitive to the current state-of-the-art approaches to sentiment and emotion classification.




New approaches for emotion classification appear every year. It is important to facilitate an easy way to benchmark and compare their performance.
For studying emotion classification with Plutchik's model, we developed a new benchmark with carefully annotated Twitter messages in three different languages. To increase the reproducibility of our work and facilitate further development in this area, we released this benchmark to the public access. We also released the RBEM-Emo patterns extracted from the training dataset.

%
%

\bibliographystyle{acl}
\bibliography{acl_rbememo}

\end{document}